\documentclass[sigconf]{acmart}

\usepackage[table]{xcolor} 
\usepackage{tabularx} 
\usepackage{booktabs}      
\usepackage{colortbl} 
 
\usepackage{amssymb}
\usepackage{pifont}
\newcommand{\cmark}{\ding{51}}
\newcommand{\xmark}{\ding{55}}

\usepackage{booktabs}
\usepackage{tabularx}
\usepackage{array}
\usepackage{makecell}

\usepackage{enumitem}
\usepackage[table]{xcolor} 
\usepackage{tabularx} 
\usepackage{booktabs}      
\usepackage{colortbl} 
\usepackage{tcolorbox}
\tcbuselibrary{skins, breakable}
 
\usepackage{amssymb}
\usepackage{pifont}

\AtBeginDocument{%
  }

\copyrightyear{2026}
\acmYear{2026}
\setcopyright{cc}
\setcctype{by}
\acmConference[MM '26]{Proceedings of the 34th ACM International Conference on Multimedia}{November 10--14, 2026}{Rio de Janeiro, Brazil}
\acmISBN{978-1-4503-XXXX-X/2018/06}

\begin{document}

\title{SI-Edit: Toward \underline{S}ketch-\underline{I}nstruction Guided Local Image Editing with Pixel-Level Precision}



\author{Weixin Ye}
\orcid{0000-0002-9837-9615}
\affiliation{%
  \institution{Institute of Information Science, \\Beijing Jiaotong University}
  \institution{Visual Intelligence + X International Joint Laboratory}
  \city{Beijing}
  \country{China}
}
\email{weixinye@bjtu.edu.cn}

\author{Wei Wang}
\orcid{0000-0002-5477-1017}
\authornote{Corresponding author.}
\affiliation{%
  \institution{Institute of Information Science,\\ Beijing Jiaotong University}
  \institution{Visual Intelligence + X International Joint Laboratory}
  \city{Beijing}
  \country{China}
}
\email{wei.wang@bjtu.edu.cn}

\author{Hongguang Zhu}
\orcid{0000-0002-1356-5153}
\affiliation{%
  \institution{City University of Macau}
  \city{Macau}
  \country{Macau}
}
\email{zhuhongguang1103@gmail.com}

\author{Xuecheng Nie}
\orcid{0000-0003-2433-5983}
\affiliation{%
 \institution{Meitu Inc}
  \city{Beijing}
  \country{China}
  }
\email{nxc@meitu.com}





\renewcommand{\shortauthors}{Weixin Ye, Wei Wang, Hongguang Zhu, and Xuecheng Nie}

\begin{abstract}
Despite rapid advances in generative models, achieving pixel-level precision in sketch-based image editing remains a persistent challenge, particularly for fine-grained local deformations. This gap stems primarily from the critical shortage of high-quality, publicly available benchmark datasets that jointly provide geometric constraints and semantic instructions. To address this issue, we first introduce \textbf{SI-Data}, a high-quality dataset specifically designed for instruction-guided local sketch editing. We develop an automated pipeline leveraging Multimodal Large Language Models (MLLMs) to synthesize comprehensive quadruplets comprising original images, local geometric sketches, semantic instructions, and corresponding edited images. By providing both reliable spatial anchors and explicit semantic intent, SI-Data uniquely enables collaborative spatial-semantic learning. Building upon this, we propose a collaborative framework called \textbf{SI-Edit} that integrates semantic instructions with precise geometric constraints. Furthermore, to address the lack of standardized evaluation, we establish a comprehensive set of metrics designed to measure both structural fidelity (e.g., sketch-to-edge alignment) and semantic adherence. Experimental results demonstrate that SI-Edit provides more reliable structural control than baselines for sketch-based image editing, and achieves precise, pixel-level local refinements aligned with user intent. The data and code are released on the \href{https://github.com/ywxsuperstar/SIEdit}{project page}.
\end{abstract}


\begin{teaserfigure}
  \includegraphics[width=\textwidth]{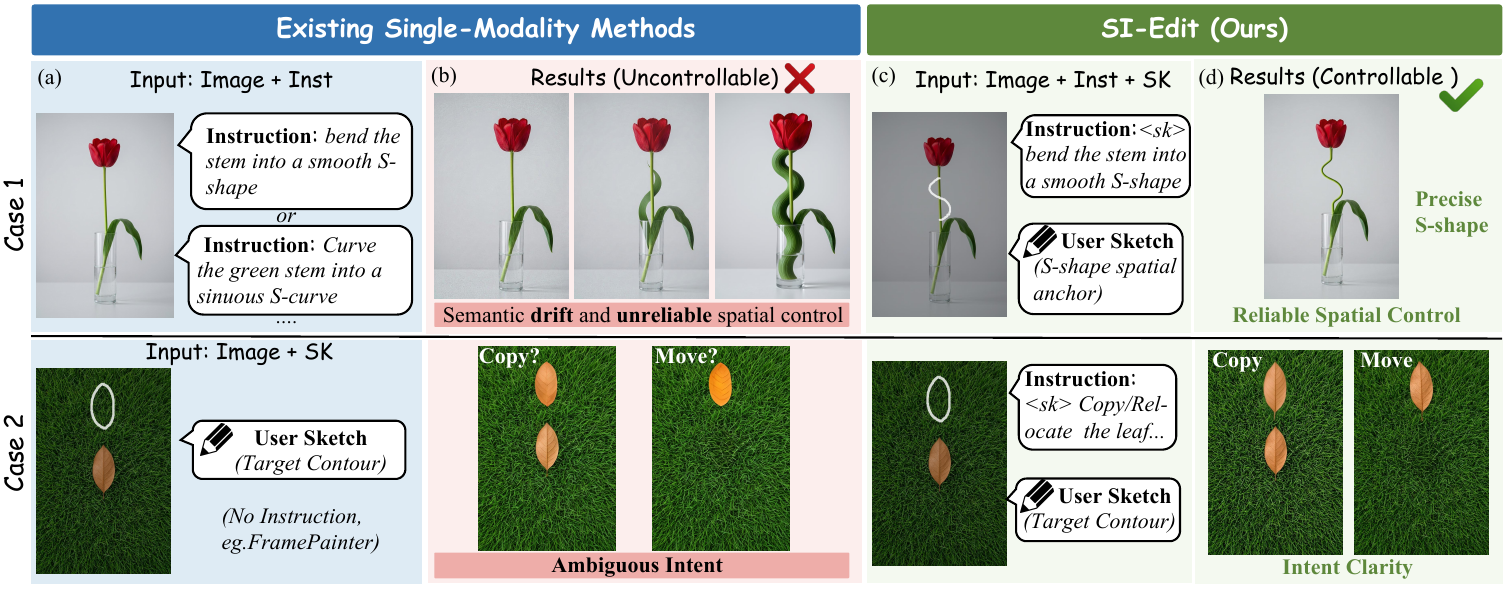}
  \caption{\textbf{Motivation.} Left (a, b): \textbf{Case 1 (Linguistic Bottleneck):} text-only instructions lack precise spatial anchors, resulting in ignored instructions, inaccurate deformation, or semantic drift; \textbf{Case 2 (Semantic Blindness):} sketch-only inputs lack explicit intent, leading to unintended operations (e.g., copy vs. move). Ours (c, d): by introducing a collaborative guidance mechanism, SI-Edit resolves both spatial and semantic ambiguities, enabling reliable, pixel-level editing aligned with user intent.}
  \label{fig:teaser}
\end{teaserfigure}



\maketitle

\section{Introduction}
The rapid advancement of generative diffusion models \cite{ho2020denoising, ramesh2022hierarchical, rombach2022high,peebles2023scalable,lipman2023flowmatchinggenerativemodeling, pmlr-v235-esser24a} has greatly improved image editing, enabling users to synthesize desired content with simple text instructions. However, as user requirements shift toward fine-grained local refinements, achieving satisfactory pixel-level precision requires a clearer expression of intent. A simple instruction like "\textit{bend the stem into a smooth S-shape}" does not specify the exact curvature or pixel-level spatial coordinates. 
Consequently, without reliable spatial anchors, such instructions often result in semantic drift and imprecise spatial control, as shown by FLUX.1 Kontext~\cite{labs2025flux1kontextflowmatching} results in Fig~\ref{fig:teaser} (\textit{Case 1}).  

To mitigate the spatial ambiguity inherent in text-only guidance, various approaches have introduced explicit spatial constraints. For instance, GLIGEN~\cite{li2023gligenopensetgroundedtexttoimage} improves spatial control by introducing a gated self-attention layer and encoding bounding boxes through Fourier embeddings. Moreover, some works, e.g., FireEdit\cite{zhou2025fireeditfinegrainedinstructionbasedimage} and MGIE\cite{fu2024mgie}, leverage Multimodal Large Language Models (MLLMs) to enhance instruction comprehension or derive regional guidance (e.g., bounding boxes), for edit localization. 
However, such region-level guidance specifies only where an edit should occur, rather than how the object's geometry should change within that region. Consequently, it remains too coarse to support precise local deformations. Furthermore, employing MLLMs as intermediate reasoning modules introduces additional computational overhead.


To provide finer-grained geometric guidance, interactive editing frameworks often introduce local sketches as explicit spatial constraints. Existing local sketch-guided editing methods mainly include inpainting-based and motion-based approaches.
Inpainting-based methods, such as SketchEdit~\cite{zeng2021sketcheditmaskfreelocalimage}, achieve mask-free editing by filling stroke-guided regions with structure-agnostic style vectors. However, relying only on geometric strokes leaves the edit intent ambiguous, making it difficult to determine whether the user expects deformation, replacement, or object addition. 
MagicQuill~\cite{liu2024magicquill} attempts to mitigate this issue by employing an MLLM to infer user intent from abstract strokes, but the inferred intent may still deviate from the desired edit.  
More critically, inpainting-based formulations tend to regenerate the target region rather than explicitly deform the original structure, limiting their ability to perform precise geometric deformations such as bending or stretching.
Motion-based methods, such as FramePainter~\cite{zhang2025framepainterendowinginteractiveimage}, reformulate image editing as an image-to-video generation task and leverage dynamic priors to reduce data requirements.
\begin{figure}[!t]
\vspace{-2mm}
  \centering
  \includegraphics[width=\linewidth]{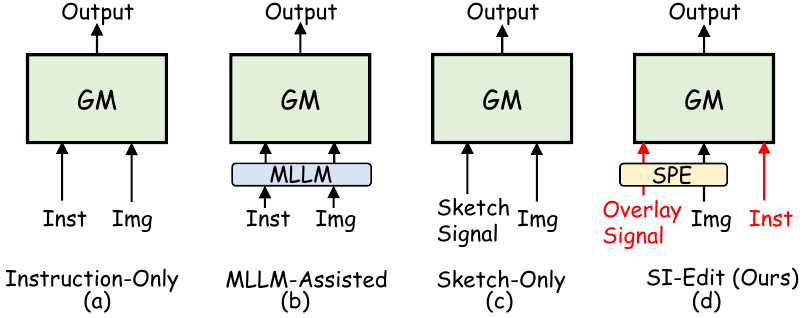}
  \caption{
    Comparison of different paradigms for image editing.
    (a) Instruction-Only methods (e.g., InstructPix2Pix~\cite{brooks2022instructpix2pix})
    suffer from spatial ambiguity.
    (b) MLLM-Assisted methods (e.g., MGIE~\cite{fu2024mgie}) improve edit
    localization but remain constrained by coarse region-level guidance and
    computational overhead.
    (c) Sketch-Only methods (e.g., SketchEdit~\cite{zeng2021sketcheditmaskfreelocalimage})
    provide geometric control but suffer from semantic blindness.
    (d) SI-Edit (Ours) combines semantic instructions and sketch constraints
    through Same Position Encoding (SPE) and an overlay signal, resolving
    spatial and semantic ambiguities for precise pixel-level editing.
    \vspace{-6mm}
  }
 \label{fig:intro-comp.}
\end{figure}
However, without explicit semantic instructions, the same sketch can correspond to different actions, such as copying or moving, leading to intent ambiguity, as illustrated in Figure~\ref{fig:teaser} (\textit{Case 2}).
Overall, although these methods provide explicit spatial grounding, they still struggle to simultaneously capture unambiguous semantic intent and preserve the original object structure during pixel-level local deformations.


Beyond algorithmic limitations, the advancement of sketch-guided editing is also hindered by a scarcity of suitable training data. Existing image-editing resources~\cite{ju2023direct,wang2023imageneteditor,kawar2023imagic,brooks2022instructpix2pix,Zhang2023MagicBrush} mainly focus on text-guided or region-guided editing, providing textual guidance or coarse masks rather than fine-grained local sketches. Such data is insufficient for supervising geometric deformations, which require sketch-level spatial signals paired with corresponding target images. 
Meanwhile, existing sketch-based methods~\cite{zeng2021sketcheditmaskfreelocalimage,zhang2025framepainterendowinginteractiveimage} rely on task-specific sketch-editing supervision, yet the corresponding paired resources are not publicly available in a standardized form, leaving a gap for a public benchmark.

Ultimately, existing methods struggle to simultaneously achieve semantic expressiveness and geometric precision for fine-grained local editing.  
More concretely, instruction-only methods provide high-level semantic descriptions but lack precise geometric guidance; region-guided methods improve localization but remain too coarse for fine-grained deformation; and sketch-only methods offer explicit geometric cues but leave the edit intent ambiguous, as illustrated in Figure~\ref{fig:intro-comp.}. 
These limitations suggest that precise local editing requires the collaborative use of both semantic instructions and geometric sketches: instructions specify \textit{what} should be changed, while sketches indicate \textit{where} and \textit{how} the change should occur.
However, developing such a framework remains challenging due to two key obstacles:
(1) \textit{the \textbf{lack of high-quality datasets} that jointly provide fine-grained sketches, semantic instructions, and target images}; 
(2) \textit{the challenge of \textbf{aligning semantic intent with geometric sketches} to enable precise pixel-level editing}.

To address these challenges, we introduce \textbf{SI-Data}, a high-quality dataset, and \textbf{SI-Edit}, a collaborative guidance framework for sketch-instruction guided local editing. 
SI-Data provides quadruplet supervision that links semantic instructions, local sketches, and target deformations, while SI-Edit learns to fuse these signals for precise local editing.
First, to address data scarcity, we design an automated synthesis pipeline that leverages MLLMs and generative teacher models to construct SI-Data, a dataset of (Source, Target, Sketch, Instruction) quadruplets tailored for structural transformations. 
Second, we introduce a collaborative spatial-semantic tuning strategy for the FLUX architecture. Specifically, we use Same Position Encoding (SPE) to align sketch inputs with source-image features, reducing spatial drift. We further introduce a learnable $\langle sk \rangle$ task-trigger token as a semantic anchor, enabling the model to jointly follow geometric sketches and semantic instructions.


The main contributions of this work are summarized as follows:
\begin{itemize}
    \item We introduce \textbf{SI-Data}, \textbf{the first dataset providing quadruplets for precise geometric deformations}, addressing the lack of data for joint spatial and semantic supervision.
    \item We propose \textbf{SI-Edit}, a collaborative framework that jointly leverages instructions and sketches via a learnable $\langle$sk$\rangle$ token and SPE strategy, resolving spatial and semantic ambiguities.
    \item Extensive experiments show that SI-Edit \textbf{achieves superior pixel-level precision and semantic alignment}, outperforming existing sketch-based baselines.  
\end{itemize}

\section{Related Work}
\subsection{Instruction-based Image Editing}
The advent of large-scale diffusion models has ushered in a new era of image editing, enabling users to intuitively modify visual content and style through natural language. 
In instruction-based approaches, textual guidance is typically integrated into generative backbones as semantic conditioning~\cite{Ye_2026_CVPR}. Works such as InstructPix2Pix~\cite{brooks2022instructpix2pix}, MagicBrush~\cite{Zhang2023MagicBrush}, and FLUX.1 Kontext~\cite{labs2025flux1kontextflowmatching} demonstrate the effectiveness of training models to follow user instructions directly. 
However, with the rapid development of generative models, simple instructions are often no longer sufficient to achieve complex, high-fidelity editing effects. Following this trajectory, recent advancements have sought to enhance instruction comprehension by integrating MLLMs or region-aware guidance. 
For instance, MGIE~\cite{fu2024mgie} leverages MLLMs to improve instruction comprehension, while FireEdit~\cite{zhou2025fireeditfinegrainedinstructionbasedimage} exploits a region-aware vision-language model to enhance region-level editing control. While these methods improve semantic interpretation, they tend to introduce additional computational overhead without resolving the core issue: abstract language alone is often insufficient to provide the controllable, pixel-level grounding required for complex geometric deformations.

\subsection{Sketch-based Image Editing}
To address the limited ability of text-only instructions to specify fine-grained spatial structure, sketches have emerged as an intuitive form of visual guidance.
Prior sketch-related methods, including SketchyGAN~\cite{chen2018sketchygan}, ControlNet~\cite{zhang2023adding}, and others~\cite{mou2024t2i,ye2023ip-adapter,qin2023unicontrol,10.1145/3588432.3591560}, have demonstrated the effectiveness of sketches and related visual cues in guiding image structure and content.
However, these methods mainly treat sketches or related visual cues as global conditions for generation rather than localized instructions that specify editing operations.
To meet the growing demand for user-driven image manipulation, several methods have explored sketch-guided local editing.
SketchRefiner~\cite{SketchRefiner} refines free-form sketches to guide interactive image inpainting, while SketchEdit~\cite{zeng2021sketcheditmaskfreelocalimage} predicts target modification regions from the input image and partial sketches without requiring explicit masks.
Nevertheless, sketch geometry alone often under-specifies the intended editing operation and target appearance.
MagicQuill~\cite{liu2024magicquill} alleviates this ambiguity by employing an MLLM to infer contextual prompts from user interactions, although this design can incur additional computational overhead.
More recently, FramePainter~\cite{zhang2025framepainterendowinginteractiveimage} leverages video diffusion priors to support complex geometric manipulations from sparse visual instructions, yet it does not explicitly model high-level editing intent.
Therefore, jointly understanding high-level editing intent and executing complex geometric manipulations remains challenging.

\subsection{Datasets for Instruction- and Sketch-Guided Image Editing}
The development of sketch-instruction guided editing has been constrained by the scarcity of datasets that jointly provide semantic intent and geometric anchors. 
Existing instruction-based datasets such as InstructPix2Pix~\cite{brooks2022instructpix2pix} and MagicBrush~\cite{Zhang2023MagicBrush} provide large-scale source--target--instruction triplets.
While InstructPix2Pix does not include explicit spatial annotations, MagicBrush contains both mask-provided and mask-free examples. However, these masks only identify editing regions rather than the desired target geometry. 
Similarly, region-based benchmarks such as EditBench~\cite{wang2023imageneteditor} incorporate spatial constraints through masks, but these masks provide only region-level localization and do not encode fine-grained geometric strokes. 
Critically, none of these datasets provides the quadruplet format required for training models to address both semantic ambiguity and spatial imprecision. 
Moreover, existing sketch-based editing methods~\cite{zeng2021sketcheditmaskfreelocalimage,zhang2025framepainterendowinginteractiveimage} have not publicly released their training datasets. A comprehensive comparison of existing datasets and methods is presented in Table~\ref{tab:comparison-full}. 
To fill this gap, we introduce SI-Data, which provides such quadruplets for joint spatial-semantic learning.
Building on SI-Data, our SI-Edit jointly leverages both modalities to address the spatial ambiguity of instruction-only methods and the semantic blindness of sketch-only approaches.

\begin{table*}[!t]
\centering
\small
\setlength{\tabcolsep}{6pt}
\caption{Comprehensive comparison of SI-Data with existing related datasets. Among the compared datasets, our SI-Data is the only one that combines fine-grained sketch constraints with semantic instructions in a quadruplet format, enabling controllable geometric deformations. \cmark~and \xmark~denote the presence or absence of each attribute.}
\vspace{-4mm}
\label{tab:comparison-full}

\begin{tabularx}{\linewidth}{
@{}
>{\centering\arraybackslash}X
>{\centering\arraybackslash}X
ccccc
>{\centering\arraybackslash}X
c
@{}
}
\toprule
\textbf{Dataset}
& \textbf{Data Format}
& \textbf{Sketch}
& \textbf{Inst.}
& \textbf{Quad.}
& \textbf{Geom.}
& \textbf{Samples}
& \textbf{Edit Mode}
& \textbf{Open} \\
\midrule

InstructPix2Pix \cite{brooks2022instructpix2pix}
& $(I_s,I_t,Inst)$
& \xmark & \cmark & \xmark & \xmark
& 313K & Global (I2I) & \cmark \\

MagicBrush \cite{Zhang2023MagicBrush}
& $(I_s,I_t,M,Inst)$
& \xmark & \cmark & \cmark & \xmark
& 10.4K & Mixed (I2I) & \cmark \\

EditBench \cite{wang2023imageneteditor}
& $(I_s,I_t,M,Inst)$
& \xmark & \cmark & \cmark & \xmark
& 240$^{\dagger}$ & Local (Inpainting) & \cmark \\

\midrule

SketchEdit \cite{zeng2021sketcheditmaskfreelocalimage}
& $(I_s,I_t,Sk)$
& \cmark & \xmark & \xmark & \cmark
& -- & Local (Inpainting) & \xmark \\

MagicQuill~\cite{liu2024magicquill}
& $(I_s,Sk)$
& \cmark & \xmark$^{\ddagger}$ & \xmark & \cmark
& 24.3K$^{\ddagger}$ & Local (Inpainting) & \xmark \\

FramePainter \cite{zhang2025framepainterendowinginteractiveimage}
& $(I_s,I_t,Sk)$
& \cmark & \xmark & \xmark & \cmark
& 20K & Local (I2V) & \xmark \\

\rowcolor[HTML]{F2F2F2}
\textbf{SI-Edit / SI-Data}
& \textbf{$(I_s,I_t,Sk,Inst)$}
& \textbf{\cmark}
& \textbf{\cmark}
& \textbf{\cmark}
& \textbf{\cmark}
& \textbf{6.5K}
& \textbf{Local (I2I)}
& \textbf{\cmark} \\
\bottomrule
\end{tabularx}
\parbox{\linewidth}{\raggedright\scriptsize
$^{\dagger}$ Evaluation-only set.
$^{\ddagger}$ MagicQuill uses MLLM-inferred prompts; 24.3K refers to the Draw\&Guess data reported in the paper.
Quad.: Quadruplet; Geom.: Geometric Deformation; Open: Public data availability. $(I_s, I_t, Sk, Inst)$: source image, target image, sketch, instruction. $M$: mask. Edit Mode: Mixed covers both global and local; I2I/I2V denote image-to-image/image-to-video.
}
 \vspace{-5mm}
\end{table*}

\begin{figure*}[!t]
  \centering
  \includegraphics[width=0.97\linewidth]{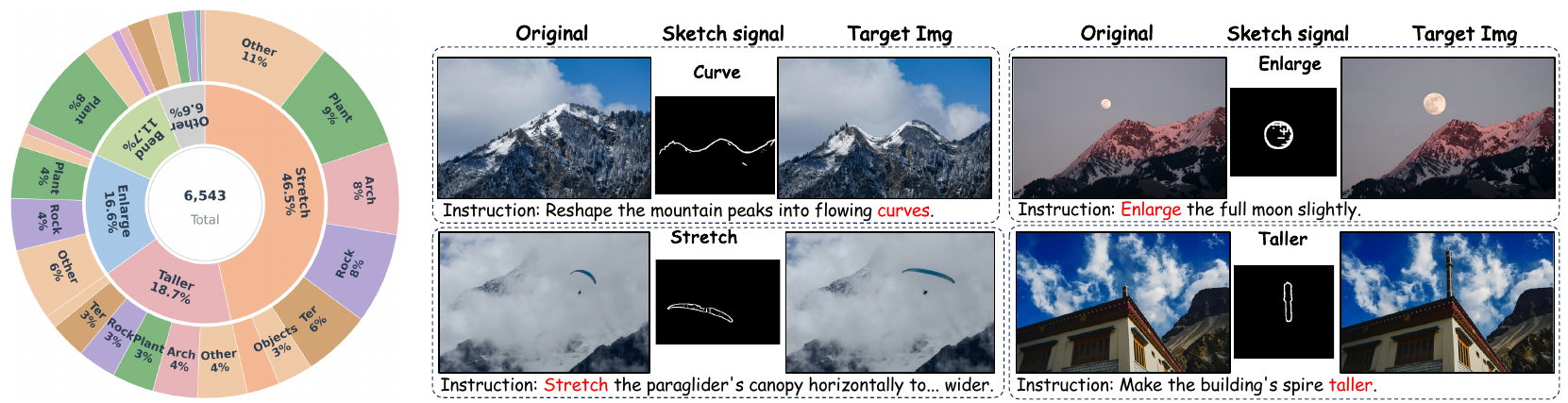}
  \vspace{-3mm}
  \caption{Overview of the constructed dataset. Left: distribution of edit types (inner ring) and target object categories (outer ring), where \textit{Arch}, \textit{Ter}, and \textit{Vehi} denote Architecture, Terrain, and Vehicles, respectively. Right: Qualitative examples from the constructed dataset, showing source images, sketch constraints, instructions, and target results.}
  \label{fig:data-distribution}
\end{figure*}

\section{SI-Data} 
\label{sec:data-construct}


\subsection{Dataset Overview}
\label{subsec:overview}
We compare SI-Data with existing image editing datasets in Table~\ref{tab:comparison-full}. 
Existing instruction-based datasets such as InstructPix2Pix~\cite{brooks2022instructpix2pix} and MagicBrush~\cite{Zhang2023MagicBrush} are organized around (source, target, instruction) triplets. MagicBrush additionally provides edit masks for mask-provided settings. Similar to the masks used in region-based benchmarks such as EditBench~\cite{wang2023imageneteditor}, these masks identify the regions to be edited but do not encode the desired geometric changes. In contrast, sketch-based methods such as SketchEdit~\cite{zeng2021sketcheditmaskfreelocalimage} and FramePainter~\cite{zhang2025framepainterendowinginteractiveimage} use sketches to provide fine-grained geometric guidance, but do not pair such guidance with semantic edit instructions. SI-Data fills this gap by providing high-quality quadruplets of (source, target, sketch, instruction).


\subsection{Dataset Construction Pipeline}
\begin{figure*}[!t]
  \centering
  \includegraphics[width=\linewidth]{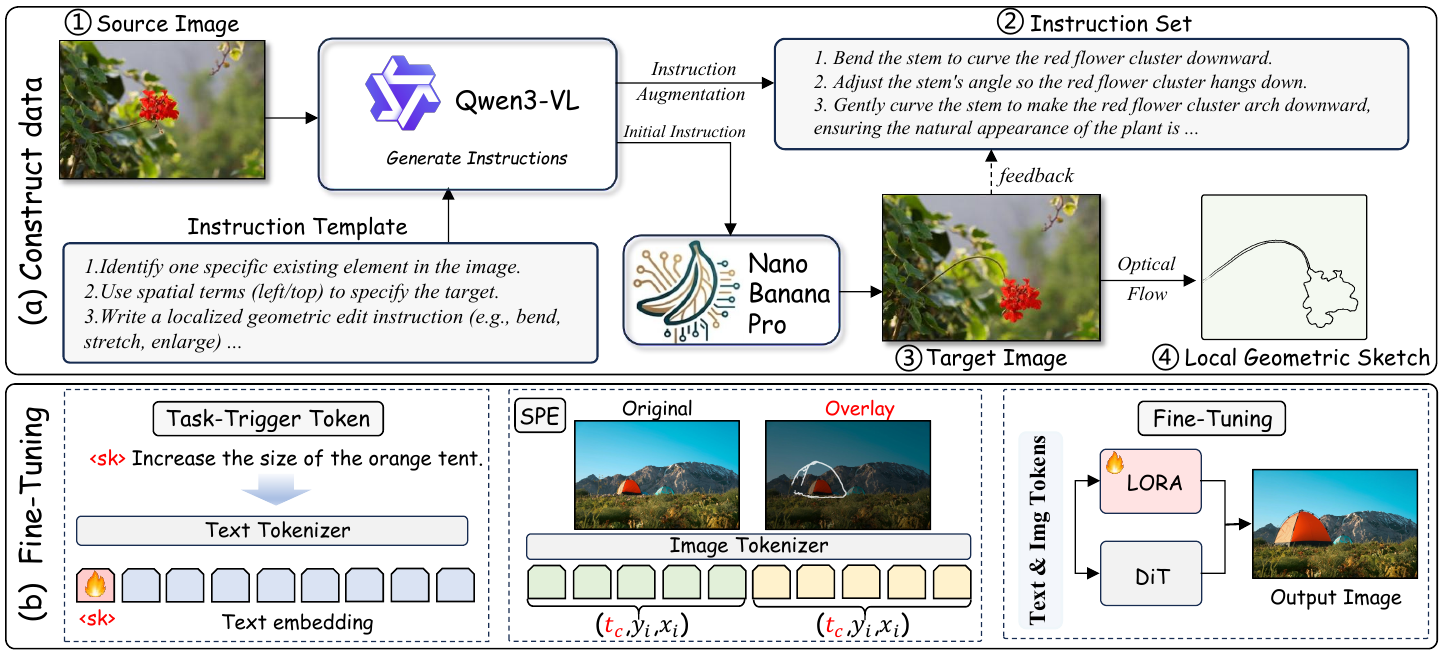}
  \vspace{-6mm}
  \caption{Overview of SI-Edit. \textbf{(a)} Data construction: Qwen3-VL generates an instruction from $I_{src}$, Nano Banana Pro produces $I_{tgt}$, and optical flow extracts the local geometric sketch $Sk$. \textbf{(b)} Method:
  Same Position Encoding (SPE) aligns the source and sketch-overlay representations, while the learnable $\langle sk \rangle$ token anchors the instruction for precise local editing.
  }
   \label{fig:construct-data} 
     \vspace{-4mm}
\end{figure*}

\leavevmode\clubpenalty=0
\label{sec:Dataset-Pipeline}
As illustrated in Figure~\ref{fig:construct-data}(a), our automatic pipeline constructs training quadruplets using high-quality Unsplash images as source images. We first prompt Qwen3-VL~\cite{qwen3technicalreport} to generate an initial spatially explicit editing instruction and then synthesize the target image using Nano Banana Pro~\cite{google2025nanobananapro}. We subsequently re-prompt Qwen3-VL with the source--target image pair and the initial instruction to construct linguistically diverse instruction variants. To obtain sparse structural constraints, we compute dense optical flow between source and target images and extract motion boundaries as binary sketches.
This construction substantially reduces manual annotation costs while enabling collaborative spatial-semantic learning. We define each data sample as a tuple $(I_{\text{src}}, I_{\text{tgt}}, Sk, Inst)$,
where $I_{\text{src}}$ and $I_{\text{tgt}}$ denote the source and target images, respectively, $Sk$ is the binary sketch extracted from optical flow, and $Inst$ is the text instruction.
Figure~\ref{fig:construct-data}(a) labels the four components as \textcircled{1} $I_{\text{src}}$, \textcircled{2} $Inst$, \textcircled{3} $I_{\text{tgt}}$, and \textcircled{4} $Sk$.
Given $(I_{\text{src}}, Sk, Inst)$, the editing task is to generate $I_{\text{tgt}}$ that satisfies three criteria: (1) semantic fidelity to $Inst$, (2) geometric alignment with $Sk$, and (3) preservation of unedited background regions. The detailed steps are as follows:

\noindent\textbf{(1) Source Image Collection.}
We use high-resolution images from the Unsplash Dataset~\cite{unsplash_dataset} as source images. In particular, we primarily use our curated \textit{Unsplash-natural} subset, whose image resolutions typically range from 2K to 6K.

\noindent\textbf{(2) Instruction Construction.}
Given a source image $I_{\text{src}}$, we prompt Qwen3-VL-30B-A3B-Instruct~\cite{qwen3technicalreport} with a structured template (see Appendix for details) to generate an initial spatially explicit editing instruction. The instruction specifies spatial localization (e.g., left or top) and a geometric operation (e.g., bend, stretch, or enlarge), while requiring all other regions to remain unchanged. This instruction is used for target image synthesis in Step~(3).

\noindent\textbf{(3) Target Image Synthesis.} 
To construct high-quality training pairs, we employ Nano Banana Pro~\cite{google2025nanobananapro} to execute the initial natural-language instruction on the source image $I_{\text{src}}$, producing the target image $I_{\text{tgt}}$.
After automatic screening followed by manual inspection, we retain $(I_{\text{src}}, I_{\text{tgt}})$ pairs that exhibit realistic, instruction-aligned geometric deformations. 

\noindent\textit{Instruction Augmentation.}
To increase linguistic diversity while maintaining consistency with the realized edit, we then re-prompt Qwen3-VL with $I_{\text{src}}$, $I_{\text{tgt}}$, and the initial instruction to generate additional semantically equivalent variants, resulting in 4--5 instructions per edit and 6.5K quadruplets in total. Each instruction in this set is denoted by $Inst$ when forming a training quadruplet. During SI-Edit training, we prepend a placeholder token $\langle sk \rangle$ to each instruction as a task trigger (see Section~\ref{sec:model-fine-tuning}).

\noindent\textbf{(4) Sketch Extraction via Optical Flow.}
For each $(I_{\text{src}}, I_{\text{tgt}})$ pair, we extract a binary sketch $Sk$ that serves as a fine-grained spatial anchor. Since the editing instructions are restricted to geometric transformations, we treat the dominant apparent displacement between $I_{\text{src}}$ and $I_{\text{tgt}}$ as arising from the intended geometric deformation.  
To obtain contour-like constraints resembling hand-drawn strokes, we use SEA-RAFT~\cite{10.1007/978-3-031-72667-5_3} to compute dense optical flow between $I_{\text{src}}$ and $I_{\text{tgt}}$.
We then extract motion boundaries by applying Sobel operators to each flow channel and thresholding the normalized gradient magnitude at 0.06. 
The resulting binary sketch $Sk$ captures motion-boundary contours associated with geometric deformation, providing sparse structural constraints.



\noindent\textbf{Dataset Statistics.}
Figure~\ref{fig:data-distribution} illustrates the statistical distribution of editing instructions generated by Qwen3-VL~\cite{qwen3technicalreport}. The dataset comprises primarily natural scenes and contains a diverse range of complex geometric transformations: \textit{Stretch/Extend} accounts for 46.5\%, followed by \textit{Taller/Higher} (18.7\%), \textit{Enlarge/Scale up} (16.6\%), \textit{Bend/Curve} (11.7\%), and \textit{Others} (6.6\%). This distribution highlights our explicit focus on fine-grained geometric deformations rather than global stylization or color modifications, providing a robust foundation for training precise geometric control.
Compared to prior work such as FramePainter~\cite{zhang2025framepainterendowinginteractiveimage}, which uses 20K video-derived training samples to learn geometric priors, our approach achieves pixel-level precision with fewer samples (6.5K) by leveraging the collaborative spatial-semantic learning paradigm introduced above.

\subsection{Evaluation Metrics}
\label{sec:Benchmark and Evaluation}
Traditional metrics, such as LPIPS~\cite{zhang2018unreasonable}, SSIM~\cite{wang2004ssim}, and global DINO~\cite{caron2021dino} features, are insufficient on their own to adequately capture the complex geometric deformations and fine-grained spatial alignments required for sketch-guided image editing. 
Specifically, image similarity metrics computed between the source and edited images may penalize intentional geometric shifts, while global feature distances may overlook pixel-level sketch adherence.
Thus, to rigorously evaluate SI-Edit, we assess three key aspects of editing performance: \textit{geometric alignment}, \textit{semantic fidelity}, and \textit{background preservation}.
Specifically, we use Localized Chamfer Distance (LCD) and Mask IoU (mIoU) for geometric alignment, CLIP score improvement ($\Delta$CS) for semantic fidelity, and Background Preservation Score (BPS) for edit-aware background preservation.

\noindent\textbf{(1) Geometric Alignment (LCD and mIoU)}
To quantify the spatial alignment between input sketch signal and the object contour extracted from the generated image, we adopt the symmetric Chamfer Distance~\cite{8099747} restricted to a localized Region of Interest (ROI). 
The ROI is defined by dilating the input sketch using the Euclidean Distance Transform, filtering out irrelevant background textures.
To extract the object contour from the generated image, we first employ SAM~\cite{kirillov2023segany} to segment the target object, using the bounding box and centroid of the input sketch as prompts.
Contour edges are then extracted from the segmentation mask obtained from SAM.
Formally, the LCD is computed as:
\begin{equation}
\text{LCD} = \frac{1}{|S_1|} \sum_{x \in S_1} \min_{y \in S_2} \|x - y\|_2 + \frac{1}{|S_2|} \sum_{y \in S_2} \min_{x \in S_1} \|y - x\|_2,
\end{equation}
where $S_1$ denotes the input sketch edge set, and $S_2$ denotes the contour edges extracted from the SAM-segmented object in the generated image within the ROI.
A lower LCD indicates superior structural adherence to the user's geometric intent.
Additionally, we compute mIoU as the intersection-over-union between dilated sketch and object contour, offering a complementary coverage-based measure of spatial alignment.

\noindent\textbf{(2) Semantic Fidelity ($\Delta$CS)}
To assess whether the edited image better aligns with the instruction, we compute the change in CLIP score relative to the source image:
\begin{equation}
\Delta \text{CS} = \text{CLIP}(I_{\text{edit}}, \text{Inst}) - \text{CLIP}(I_{\text{src}}, \text{Inst}),
\end{equation}
where $I_{\text{src}}$ and $I_{\text{edit}}$ denote the source and edited images, respectively, $\text{Inst}$ denotes the editing instruction, and $\text{CLIP}(\cdot,\cdot)$ denotes the cosine similarity between normalized image and text embeddings.
A positive $\Delta$CS indicates increased CLIP-based alignment with the editing instruction, and a higher $\Delta$CS suggests greater semantic improvement.

\noindent\textbf{(3) Background Preservation Score (BPS).}
A background-only metric may assign high scores to outputs that remain close to the input but fail to perform the requested edit. 
We therefore propose BPS to jointly evaluate \textit{background preservation} and \textit{edit fidelity}. 
Let $I_{\text{tgt}}$ denote the ground-truth target image.
We define the expected edit and background regions as
\begin{equation}
M_{\mathrm{edit}}(p){=} \mathbb{I}\!\left(
\left\|I_{\mathrm{tgt}}(p)-I_{\text{src}}(p)\right\|_{\infty}{>}\tau \right),
\quad
M_{\mathrm{bg}}(p){=}1-M_{\mathrm{edit}}(p).
\end{equation}
$\|\cdot\|_{\infty}$ denotes the maximum absolute difference across the RGB channels. Pixels for which the difference between the source and target images exceeds a small threshold $\tau$ form the expected edit region, while the remainder form the background.
We then compute
\begin{equation}
S_{\mathrm{bg}} = \operatorname{SSIM}_{M_{\mathrm{bg}}}({I_{\text{edit}}},I_{\text{src}}),
\quad
S_{\mathrm{edit}} = \operatorname{SSIM}_{M_{\mathrm{edit}}}
({I_{\text{edit}}},I_{\mathrm{tgt}}),
\end{equation}
where $\operatorname{SSIM}_{M}(A,B)$ denotes the mean of the pixel-wise SSIM map between $A$ and $B$ over region $M$. Finally, BPS is defined as the harmonic mean of the two scores:
\begin{equation}
\mathrm{BPS} = \frac{2S_{\mathrm{bg}}S_{\mathrm{edit}}} {S_{\mathrm{bg}}+S_{\mathrm{edit}}}.
\end{equation}
A high BPS requires the model to preserve regions that should remain unchanged while accurately reconstructing regions that should be edited. Consequently, BPS penalizes both under-editing and unintended background changes.


\section{Method}
We introduce SI-Edit, a framework for precise, fine-grained image editing under sketch guidance, built on the constructed SI-Data (Sec.~\ref{sec:data-construct}). SI-Edit resolves the dual challenges of semantic blindness (sketch-only) and spatial ambiguity (instruction-only) through a collaborative spatial-semantic fine-tuning strategy. 
This approach leverages same position encoding and a learnable $\langle sk \rangle$ task anchor within a Multi-Modal Diffusion Transformer, effectively aligning abstract linguistic intents with fine-grained geometric anchors.


\subsection{Preliminaries}
We adopt FLUX.1 Kontext~\cite{labs2025flux1kontextflowmatching}, a latent Diffusion Transformer (DiT)~\cite{peebles2023scalable}, as our backbone. Following rectified flow~\cite{liu2022flow}, a variant of flow matching~\cite{lipman2023flowmatchinggenerativemodeling}, the model is trained to learn a velocity field $v_t$ that transports samples from a noise distribution to the target data distribution.
The velocity field $v_t$ is parameterized by a Multi-Modal DiT (MM-DiT) backbone~\cite{pmlr-v235-esser24a, peebles2023scalable}. 
Its Double Stream Blocks retain modality-specific transformations while enabling joint attention between text and image tokens, whereas the subsequent Single Stream Blocks operate on their concatenated representations. Spatial information is encoded using positional embeddings
~\cite{ye2025unified,min2026compassrope}, with FLUX.1 Kontext specifically adopting 3D Rotary Positional Embeddings (RoPE)~\cite{labs2025flux1kontextflowmatching}.
This spatial grounding, combined with modality-specific processing during the early stages, establishes the architectural foundation for decoupling abstract semantic intent from precise, fine-grained geometric control.

\subsection{Collaborative Spatial-Semantic Learning}
\label{sec:model-fine-tuning}
To align semantic intent with geometric anchors, we propose a collaborative spatial-semantic strategy within the DiT architecture. As illustrated in Figure~\ref{fig:construct-data} (b), this strategy integrates instruction and sketch signals into a joint latent space for precise alignment. 
We introduce a task trigger token $\langle sk \rangle$ and SPE for this purpose.

\noindent\textbf{Task-Trigger Token $\langle sk \rangle$.}
Unlike conventional instruction-based edits, in which the intent is conveyed solely through text embeddings from encoders such as T5-XXL~\cite{colin2020exploring} or CLIP~\cite{radford2021learningtransferablevisualmodels}, we introduce a specialized task-trigger token $\langle sk \rangle$ for sketch-guided geometric editing.
Specifically, for each text encoder, we add the $\langle sk \rangle$ token to its vocabulary and separately initialize its embedding using the mean embeddings of the tokens in the phrase \textit{``follow the sketch lines''}.
\begin{figure*}[!t]
  \centering
  \includegraphics[width=1.0\linewidth]{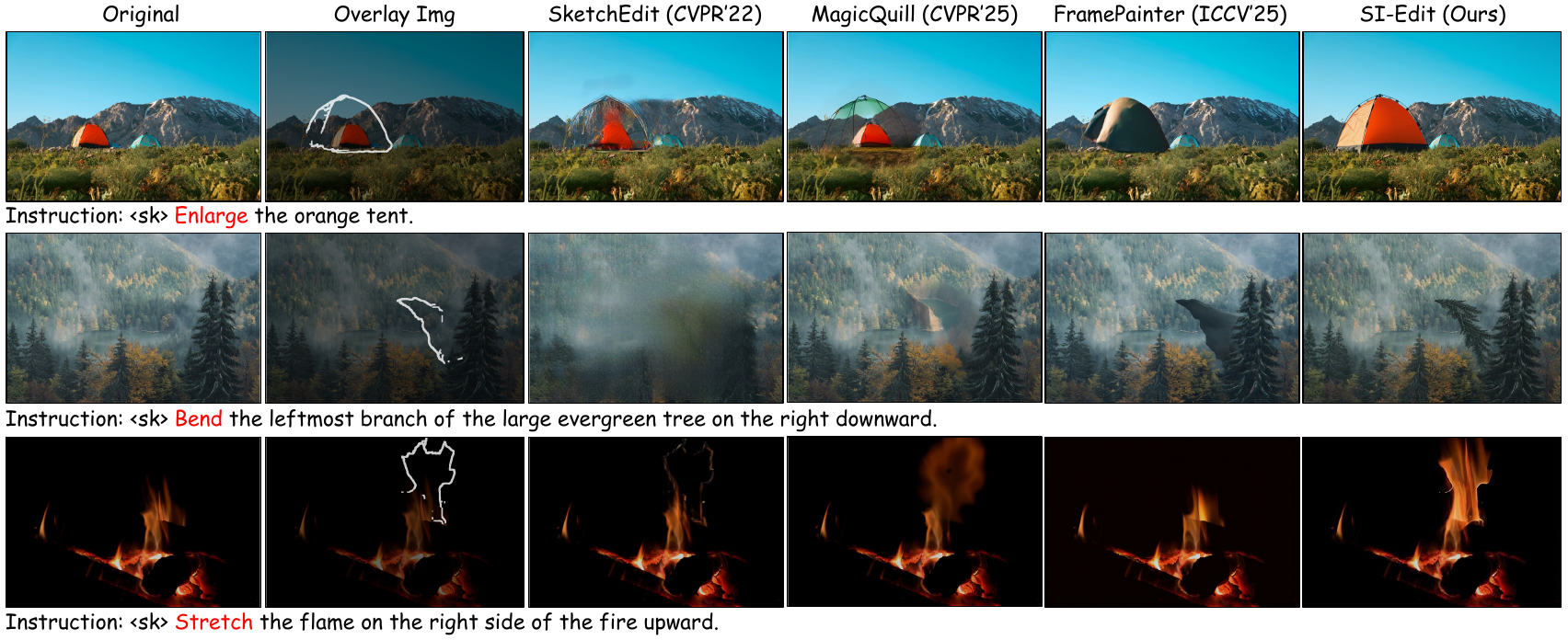}
  \vspace{-7mm}
  \caption{
Qualitative comparison of SI-Edit against baseline methods. Across \textit{Enlarge}, \textit{Bend}, and \textit{Stretch}, our method closely follows the input geometric constraints while synthesizing realistic textures that blend naturally with the unedited background.}  
   \label{fig:qualitative-res}
   \vspace{-3mm}
\end{figure*}
We then prefix each instruction with the $\langle sk \rangle$ token:
\begin{equation}
Inst = [\langle sk \rangle, T_1, \dots, T_n],
\end{equation}
where $T_1,\dots,T_n$ denote the tokens of the original instruction.
During the forward pass of the MM-DiT, the $\langle sk \rangle$ embedding acts as a global semantic anchor, conditioning the attention mechanisms across both text and image streams, and steering the generation process toward sketch-guided editing.

\noindent\textbf{Same Position Encoding (SPE) Strategy.} 
A fundamental challenge in sketch-guided editing is preserving the structural identity of the source image while executing geometric deformations. 
In standard architectures, sketch and source information are processed as separate conditioning inputs without an explicit mechanism to enforce spatial correspondence.  
This can lead to spatial misalignment, as the model may not guarantee that a sketch stroke corresponds to the same location as the intended deformation region. 
To address this, we first overlay the binary sketch $Sk$ onto the source image $I_{src}$ to obtain an overlay image $I_{overlay}$. 
We then propose the SPE strategy, where the source and overlay latents $z_{src}, z_{overlay} \in \mathbb{R}^{h \times w \times c}$ are independently patchified and projected to token sequences $T_{src}, T_{overlay} \in \mathbb{R}^{N \times d}$, where $N = hw/p^2$. We then concatenate them along the token dimension as:
\begin{equation}
T_{cond} = \text{Concat}(T_{src}, T_{overlay}) \in \mathbb{R}^{2N \times d}.
\end{equation}
$T_{\mathrm{src}}$ and $T_{\mathrm{overlay}}$ are assigned identical 3D RoPE coordinates $(t_c, y_i, x_i)$, where the two conditioning streams ($T_{\mathrm{src}}$ and $T_{\mathrm{overlay}}$) share the same conditioning index $t_c$ along the first RoPE dimension, as well as the same spatial position $(y_i, x_i)$.
Following the native Kontext conditioning mechanism, $T_{\mathrm{cond}}$ is appended to the noisy target-latent tokens and fed into the visual stream of the MM-DiT. 
This shared positional encoding facilitates spatial correspondence at the latent-token level between the source and overlay tokens, encouraging the generated edit to follow the sketch geometry while mitigating spatial drift, without requiring an additional alignment module.

\noindent\textbf{Collaborative Fine-tuning via LoRA.}
We apply Low-Rank Adaptation (LoRA)~\cite{hu2021loralowrankadaptationlarge} to the attention layers in both the Double Stream and Single Stream Blocks of the FLUX model.
Together with the learnable task-trigger embeddings, these adapters facilitate interactions between the task-conditioned text representations and the SPE-aligned visual conditions.
The text instruction provides semantic editing intent, while the source and sketch-overlay representations provide spatial guidance for the geometric transformation.
The model is optimized using the standard rectified-flow matching objective, which minimizes the discrepancy between the predicted and target velocity fields along the interpolation path.


\section{Experiments}
\subsection{Setup}
\leavevmode\clubpenalty=0
\noindent\textbf{Implementation Details.} We implement SI-Edit based on the FLUX.1 Kontext~\cite{labs2025flux1kontextflowmatching}. Following the 
setup of OminiKontext~\cite{omini-kontext}, we employ LoRA~\cite{hu2021loralowrankadaptationlarge} to fine-tune the attention layers of both the Double- and Single-Stream Blocks. We set the LoRA rank to $r{=}256$ to ensure sufficient capacity for capturing complex geometric transformations. The model is trained on a single NVIDIA H20 GPU. We utilize the Prodigy optimizer ($lr{=}1.0$) for adaptive learning rate scaling. 
During training, sketches are integrated as 0.8-opacity overlays with online augmentation to enhance geometric robustness.

\noindent\textbf{Evaluation Setup.}
Unlike prior evaluation settings~\cite{ju2023direct, wang2023imageneteditor, kawar2023imagic}, which are primarily based on region-level masks or text instructions, our evaluation leverages the quadruplet annotations (source, target, sketch, instruction) from SI-Data, enabling a finer-grained assessment of pixel-level geometric edits across 159 test samples randomly selected at the source-image level. 
Additionally, we adopt the metrics described in Sec.~\ref{sec:Benchmark and Evaluation}, namely LCD, mIoU, $\Delta$CS, and BPS, to evaluate editing performance from complementary perspectives.

\subsection{Sketch-based Geometric Image Editing}
\subsubsection{Qualitative Results}
As shown in Figure~\ref{fig:qualitative-res}, we visually compare SI-Edit against existing baseline methods on complex geometric edits, including scaling (tent), bending (tree branch), and stretching (flame). SketchEdit~\cite{zeng2021sketcheditmaskfreelocalimage} frequently struggles with ``semantic blindness'', failing to preserve the original textures and instead producing unintended and blurry artifacts within the sketched regions. 
\begin{table}[!t]
\centering
\caption{Performance comparison of existing sketch-based editing methods on various metrics.}
\vspace{-3mm}
\label{tab:main-tab}
\begin{tabularx}{\linewidth}{lXXcX} 
\toprule
\textbf{Method} & \textbf{LCD$\downarrow$} & \textbf{mIoU$\uparrow$} & $\Delta$\textbf{CS$\uparrow$} & \textbf{BPS$\uparrow$} \\
\midrule
SketchEdit \cite{zeng2021sketcheditmaskfreelocalimage} & 10.957 & 0.082 & 0.0109 &0.544 \\
MagicQuill \cite{liu2024magicquill} & \underline{7.434} &  \underline{0.231} & 0.0068 &0.576 \\
FramePainter \cite{zhang2025framepainterendowinginteractiveimage} & 7.979 & 0.129 &  \underline{0.0114} &  \underline{0.587} \\
\rowcolor[HTML]{EFEFEF}
SI-Edit (Ours) & \textbf{4.292} & \textbf{0.362} & \textbf{0.0174} & \textbf{0.648} \\
\bottomrule
\end{tabularx}
\vspace{-3mm}
\end{table}
MagicQuill~\cite{liu2024magicquill} uses brush strokes for structural guidance but relies on inpainting-based regeneration; the results for the tent, branch, and flame examples deviate from the intended contours and do not consistently preserve the original object structure.
FramePainter~\cite{zhang2025framepainterendowinginteractiveimage}, while attempting to follow the geometric constraints, often hallucinates incorrect textures (e.g., replacing the orange tent with a gray one) or creates unnatural boundaries. In contrast, SI-Edit produces edits that are semantically consistent with the instruction and spatially aligned with the sketch. 
\begin{figure*}[!t]
  \centering
  \includegraphics[width=0.99\linewidth]{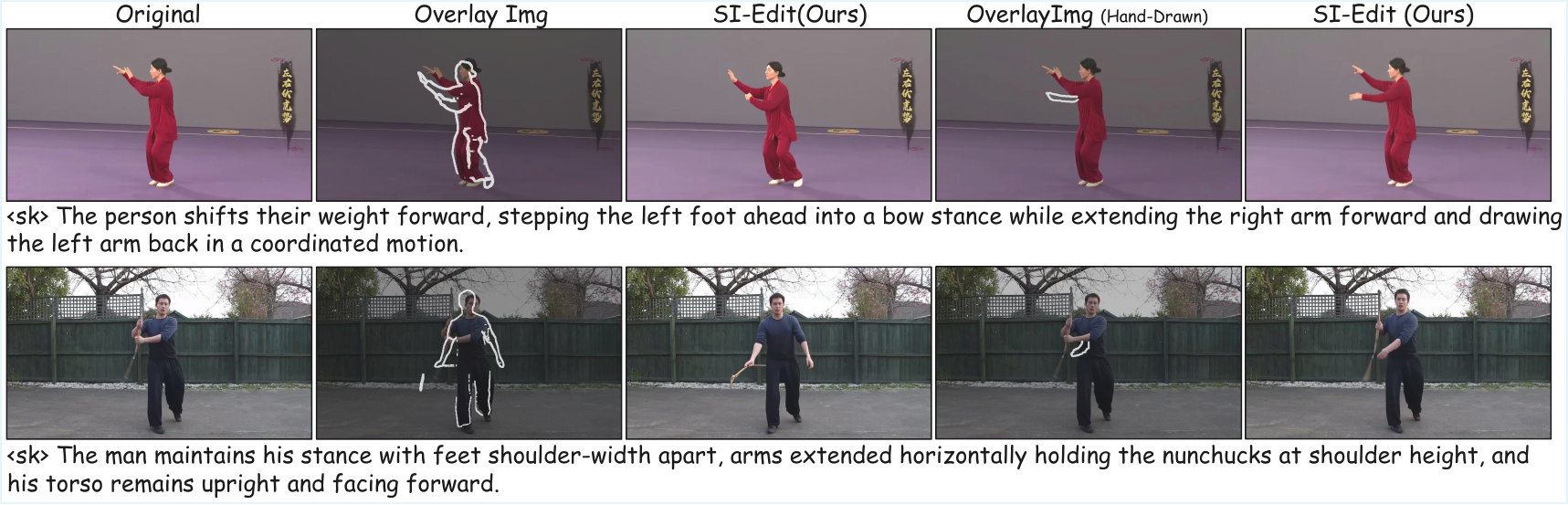}
  \vspace{-3mm}
  \caption{
  SI-Edit on sketch-guided motion editing. Given a source pose and a target-pose sketch, our method synthesizes an edited image in the target pose. Results with hand-drawn sketches further demonstrate its ability to generate body parts aligned with user-provided strokes while preserving subject appearance. Instructions: Col. 3; see the supplement for Col. 5.
  }
   \label{fig:kungfu}
   \vspace{-1mm}
\end{figure*} 
It exactly enlarges the tent, naturally bends the tree branch without disrupting the background mist, and faithfully stretches the flame along the specified contour. This demonstrates our model's superior ability to execute localized, pixel-level deformations while seamlessly preserving object identity and the appearance of unedited regions.


\subsubsection{Quantitative Results}
We present a quantitative comparison between SI-Edit and SOTA sketch-based editing baselines in Table~\ref{tab:main-tab}.
Our method achieves the best performance among all competing approaches across all evaluation metrics. For geometric fidelity, SI-Edit achieves an LCD of 4.292, substantially lower than SketchEdit (10.957), FramePainter (7.979), and MagicQuill(7.434), while its mIoU score of 0.362 is nearly three times that of FramePainter (0.129), indicating improved alignment with the input sketch and more precise localization of the edited region. 
On semantic alignment, our method yields the highest $\Delta$CS (0.0174), indicating that SI-Edit effectively captures complex linguistic intent under strict spatial constraints.
For edit-aware background preservation, SI-Edit achieves the best BPS (0.648), outperforming FramePainter (0.587), MagicQuill (0.576), and SketchEdit (0.544). This result indicates that SI-Edit is the most effective among the compared methods at faithfully performing the intended edit while preserving regions that should remain unchanged.
Collectively, these results suggest that SI-Edit effectively addresses the challenge of aligning spatial constraints with semantic intent, enabling precise pixel-level editing while outperforming existing sketch-based editing baselines.

\begin{table}[!t]
  \centering
  \caption{Ablation study on task-trigger token configuration.}
  \vspace{-3mm}
  \begin{tabular}{lccccc}
    \toprule
    \textbf{Trigger Type} & \textbf{Position} & \textbf{LCD$\downarrow$} & \textbf{mIoU$\uparrow$} & $\Delta$\textbf{CS$\uparrow$} & \textbf{BPS$\uparrow$} \\
    \midrule
    Fixed Prompt & BOS & \textbf{4.023} & 0.332 & \underline{0.0149} & \textbf{0.669} \\
    Learnable $\langle sk \rangle$ & EOS & 4.352 & \underline{0.358} & 0.0052 & 0.646 \\
    \rowcolor[HTML]{EFEFEF}
    Learnable $\langle sk \rangle$ & BOS & \underline{4.292} & \textbf{0.362} & \textbf{0.0174} & \underline{0.648} \\
    \bottomrule
  \end{tabular}
  \label{tab:sk-ablation}
  \vspace{-4mm}
\end{table}

\subsection{Sketch-Guided Movement Editing}
\leavevmode\clubpenalty=0
Beyond object-level deformations, our method can be extended to movement editing—altering a subject's pose (e.g., changing a character's stance or moving individual limbs) while preserving identity and appearance. Unlike static manipulation, it requires reasoning about large, structurally uncertain regions from only sparse sketches while handling occlusions appropriately. We evaluate on the \textit{kungfu} subset of Motion-X~\cite{NEURIPS2023_4f8e27f6}, which provides diverse human motion sequences. 
For each test, we extract source-target frame pairs from distinct actions, then follow the pipeline in Sec.~\ref{sec:Dataset-Pipeline} to obtain a geometric sketch of the motion trajectory and a semantic instruction describing the target pose.
The result is shown in Figure~\ref{fig:kungfu}, demonstrating two scenarios. With automatically extracted sketches, SI-Edit well performs the full pose transition (left foot forward, arms adjusted) in Figure~\ref{fig:kungfu} Col.3. With a rough hand-drawn stroke indicating the desired arm movement, the left arm is successfully moved downward as instructed while preserving the standing pose, right arm position, and overall appearance (in Figure~\ref{fig:kungfu} Col.5).    
These results confirm our method's effectiveness for movement editing and its robustness to both automatically extracted sketches and flexible hand-drawn inputs.

\subsection{Additional Experiments}
\noindent\textbf{Effect of Task-Trigger Token Configuration.}
To investigate the impact of the task trigger $\langle sk \rangle$, we compare the learnable $\langle sk \rangle$ token against a fixed prompt initialized with the mean embedding of the prompt \textit{``following the sketch lines''}. 
As reported in Table~\ref{tab:sk-ablation}, the learnable token at Beginning-of-Sequence (BOS) improves semantic alignment and achieves a higher mIoU, with $\Delta$CS increasing from 0.0149 to 0.0174 and mIoU from 0.332 to 0.362.
The fixed prompt yields better LCD (4.023 vs.\ 4.292) and BPS (0.669 vs.\ 0.648), revealing a trade-off across the evaluation metrics.
Furthermore, we evaluate the token's position sensitivity by placing $\langle sk \rangle$ at the BOS versus End-of-Sequence (EOS). 
Placing $\langle sk \rangle$ at the BOS substantially improves semantic alignment ($\Delta$CS: 0.0174 vs.\ 0.0052), while the other metrics remain comparable. 
These results suggest that placing $\langle sk \rangle$ before the instruction, rather than after it, provides more effective task conditioning for sketch-guided image editing.


\noindent\textbf{User Study.}
We conduct a pairwise user study on 60 test cases randomly selected from the 159-sample test set. Participants compare SI-Edit with each baseline in terms of visual consistency, editing accuracy, and overall image quality. For each comparison, participants are shown two anonymized editing results presented in a randomized left--right order and select the preferred result separately for each of the three criteria. We obtain 620 valid pairwise comparison responses, each containing three criterion-specific preference judgments. 
Table~\ref{tab:user-study} reports, for each criterion, the percentage of valid comparisons in which SI-Edit is preferred. SI-Edit receives higher preference rates than all three baselines across all three criteria. 
\begin{table}[!t]
\centering
\caption{User preference rates for SI-Edit over each baseline across three evaluation criteria (\%).}
\vspace{-3mm}
\label{tab:user-study}
\begin{tabularx}{\linewidth}{
    @{}l
    *{3}{>{\centering\arraybackslash}X}
    @{}
}
\toprule
\textbf{Method}
& \textbf{Visual Cons.}
& \textbf{Edit Acc.}
& \textbf{Image Qual.} \\
\midrule
MagicQuill   & 81.3 & 83.3 & 82.3 \\
SketchEdit   & 91.9 & 93.3 & 94.8 \\
FramePainter & 82.6 & 83.1 & 87.4 \\
\bottomrule
\end{tabularx}
\vspace{-4mm}
\end{table}

\section{Conclusion}
Progress in sketch-guided image editing is hindered by the limited availability of datasets that jointly provide semantic instructions and fine-grained geometric sketches. To address this gap, we introduce SI-Data, a quadruplet dataset constructed through an MLLM-assisted pipeline. Building on SI-Data, we propose SI-Edit, a collaborative guidance framework that integrates semantic intent with geometric anchors through a learnable $\langle sk \rangle$ token and the SPE strategy. Our experiments show that SI-Edit outperforms the compared sketch-based baselines and supports pixel-level control over local geometric edits. Together, SI-Data and the accompanying evaluation protocol provide a transparent benchmark intended to facilitate research on sketch-instruction guided image editing.
\bibliographystyle{ACM-Reference-Format}
\balance
\bibliography{samples/sample-base}

\clearpage
\appendix
\section{Implementation Details}
\subsection{Training Details} 
We mainly build upon the official implementation of OminiKontext~\cite{omini-kontext}, which is based on FLUX.1 Kontext~\cite{labs2025flux1kontextflowmatching}. The model is trained in bfloat16 to optimize memory efficiency while maintaining numerical stability. We train the model for up to 10,000 steps. For the LoRA configuration, we set the scaling parameter $\alpha=256$ to match the rank $r=256$. During optimization with Prodigy (with the default learning rate of 1), we apply a weight decay of 0.01 and enable both bias correction and safeguard warmup. To enable classifier-free guidance during inference, we randomly drop the text conditioning with a probability of 0.1. 

\subsection{Instruction Template} 
We employ the Qwen3-VL-30B-A3B-Instruct~\cite{qwen3technicalreport} VLM to automatically generate localized geometric editing instructions for our training dataset. To ensure high-quality and precise text-image alignment, we design a carefully crafted instruction template. We prompt the model to identify a single element within the input image and propose a localized spatial modification (e.g., \textit{bending}, \textit{enlarging}, or \textit{stretching}) while strictly maintaining the rest of the scenery unchanged. To guide the model's behavior, the template includes both positive examples (e.g., ``\textit{Bend the plant stem into a graceful curve}'') and negative examples of overly broad edits (e.g.,``\textit{Transform all mountain peaks}''). Furthermore, we enforce the use of spatial specifiers (e.g., ``\textit{leftmost}'', ``\textit{foreground}'') to disambiguate target selection when multiple similar objects are present. To facilitate automated pipeline integration, the VLM is constrained to output the results strictly as a JSON object containing three fields: the target subject, the spatial action, and the final editing instruction.
The pixel-level sketch is then extracted separately from the resulting source--target image pair using optical flow. Together, the source image, target image, extracted sketch, and generated instruction are then assembled into a training quadruplet, providing complementary semantic intent and fine-grained spatial guidance for local geometric editing.




\section{Prompt Information}
In this section, we provide the prompts used by Nano Banana Pro to generate source images shown in the main paper and appendix. All corresponding editing results are produced by our SI-Edit.

\subsection{Main Paper}
\noindent\textbf{Figure 1- case1:} \textit{A minimalist and elegant photorealistic image of a single vibrant red tulip. The tulip has a long, straight green stem with one large green leaf, placed in a simple, clear glass vase half-filled with water. The vase sits on a clean, flat surface against a soft, plain light-grey background. Centered composition, soft diffused studio lighting, crisp details, 8k resolution, highly detailed, extremely sharp focus, professional product photography.} 

\noindent\textbf{Figure 1- case2:} \textit{An ultra-sharp, high-resolution top-down photo of a meticulous green lawn with dense, detailed blades. At the absolute center, a single, solitary, smooth-edged oval brownish-orange fallen leaf is precisely positioned. The leaf features a simple, non-serrated lanceolate shape with clearly visible, intricate vein structures. The leaf occupies approximately one-sixth of the vertical frame. Soft natural lighting, 8k resolution, photorealistic, and minimalist composition.}

\begin{tcolorbox}[colback=gray!5!white, colframe=gray!60!black, title=\textbf{Instruction Generation Prompt Template}, arc=3mm, width=\linewidth, left=2mm, right=2mm, top=2mm, bottom=2mm, breakable]
\small
Create an instruction for LOCALIZED geometric editing of ONE specific part in the image.\\

\textbf{TASK:}\\
Find ONE element that ACTUALLY EXISTS in the image, and write a clear instruction to modify ONLY that element. All other elements and scenery must remain UNCHANGED.\\

\textbf{GOOD EXAMPLES:}\\
- ``Bend the plant stem into a graceful curve, making the flower cluster arch downward''\\
- ``Enlarge/Expand the leftmost petal in the foreground''\\
- ``Stretch the building's spire upward''\\
- ``Elongate the center hydrangea cluster into a cascading form'' (specifies WHICH one among multiple similar elements)\\
- ``Add a small petal next to the flower''\\

\textbf{BAD EXAMPLES:}\\
- ``Transform all mountain peaks'' (too broad)\\
- ``Stretch the whole building taller'' (changes entire object)\\
- ``Bend the tree trunk, making all branches twist'' (affects multiple parts)\\
- ``Reshape the hydrangea flower clusters into elongated forms'' (should specify which one)\\

\textbf{REQUIREMENTS:}\\
1. Only describe what you ACTUALLY see (if you see a building, say ``building'', not ``tree'')\\
2. Choose ONE part. When there are MULTIPLE similar elements (e.g., several flowers, multiple trees, many rocks), specify which one using position words like: ``the left/right/center one'', ``the top/bottom one'', ``the foreground/background one'', ``the largest/smallest one''.\\
3. Write an instruction to modify that part:\\
\hspace*{4mm} - ``Bend/Deform [the part] into a curve/shape/position''\\
\hspace*{4mm} - ``Enlarge/Expand [the part]''\\
\hspace*{4mm} - ``Stretch/Move [the part] horizontally/vertically/left/right/up /down''\\
\hspace*{4mm} - ``Add a small [element]''\\

\textbf{Output Format (JSON only):}\\
\texttt{\{}\\
\texttt{  "target\_subject": "the part (e.g., 'plant stem', 'petal', 'one branch')",}\\
\texttt{  "spatial\_action": "the action (e.g., 'bending', 'resizing', 'stretching')",}\\
\texttt{  "instruction": "instruction (e.g., 'Bend the plant stem into a graceful curve')"}\\
\texttt{\}}\\

\textbf{RULES:}\\
- Only edit ONE part\\
- Only describe elements that EXIST in the image\\
- Keep instructions simple and clear\\
- Output ONLY the JSON string.
\end{tcolorbox}

\subsection{Additional Instructions}
The following instructions, paired with the hand-drawn sketches, are used to generate the fifth-column results shown in Figure~6 of the main paper.

\noindent\textbf{(Top)} $\langle sk \rangle$ \textit{The woman moves her left arm downward, while maintaining her original standing pose and right arm position}.

\noindent\textbf{(Bottom)} $\langle sk \rangle$ \textit{While strictly maintaining his current stance and right arm pose, the man shifts his left arm downwards}.


\begin{figure*}[!t]
  \centering
  \includegraphics[width=\linewidth]{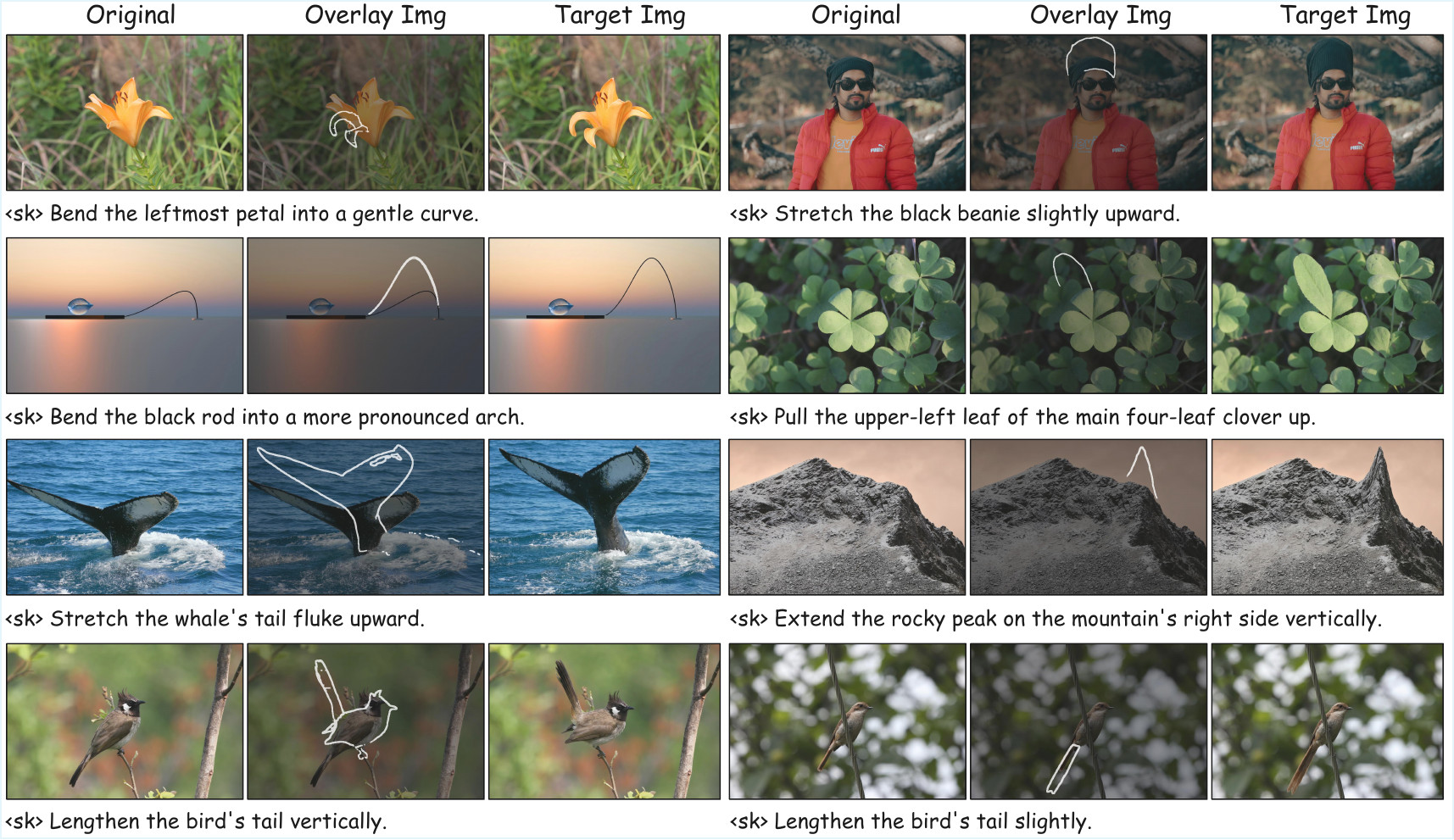}
  \caption{Some examples from our constructed dataset. For each case, we display the original image (left), the sketch condition overlaid on the original image (middle), and the target image (right). The text below each example is the VLM-generated localized editing instruction.}  
 \label{fig:sup-data-show}
 \vspace{-2mm}
\end{figure*}

\begin{figure*}[!t]
  \centering
  \includegraphics[width=\linewidth]{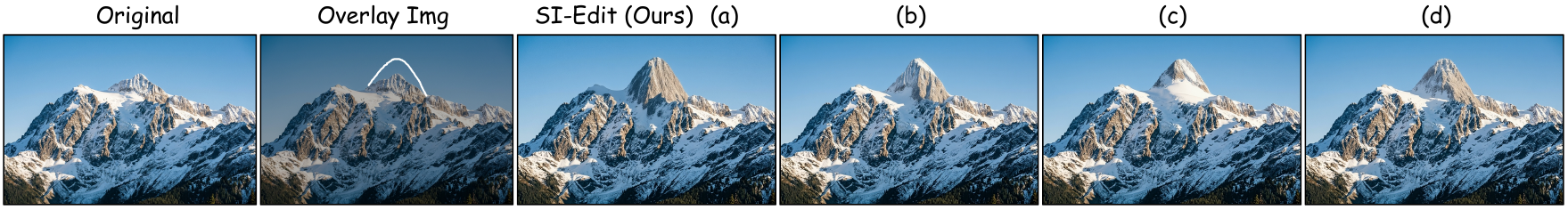}
  \caption{Qualitative results on robustness to diverse textual prompts. Columns (a)-(d) correspond to the four prompts listed in Sec.~\ref{sec:sup-prompt-generation}. Our model consistently follows the structural guidance of the sketch regardless of prompt complexity.}  
  \vspace{-2mm}
 \label{fig:sup-generation-prompt}
\end{figure*}

\subsection{Appendix Content}

\noindent\textbf{Figure}~\ref{fig:sup-generation-prompt}: \textit{A majestic snow-capped mountain peak under a clear sky. Prominent rocky ridges partially covered in pristine white snow. Photorealistic, 8K resolution, stunning landscape photography, sharp focus, natural lighting, high contrast.}

\noindent\textbf{Figure}~\ref{fig:sup-generation-sketch}: \textit{A vibrant pink water lily in full bloom, rising above dark still water on a slender brown stem. Surrounded by large, textured green and dark purplish-brown lily pads with visible veins. Macro photography, natural soft daylight, crisp focus on the pink petals, gentle water reflection, soft background blur (bokeh). 8K resolution, ultra-detailed, photorealistic, masterpiece, high contrast.}

\section{Video-based Data Construction}
To extend our method to human movement editing, we construct a video-derived training dataset from the \textit{kungfu} subset of Motion-X ~\cite{NEURIPS2023_4f8e27f6}. The pipeline consists of the following steps:

\textbf{(1) Frame Extraction.} 
We extract frames from source videos at a fixed rate of 10 FPS and limit the maximum frames per video to 100 for computational efficiency.

\textbf{(2) Frame Pair Selection.}
For each video, we identify valid frame pairs $(I_{\text{src}}, I_{\text{tgt}})$ that exhibit meaningful moving changes while maintaining identity consistency. Specifically, we compute the centroid displacement between human masks across frames and filter pairs based on:
\begin{itemize}
    \item displacement magnitude within $[d_{\min}, d_{\max}]$ pixels;
    \item area ratio change below a threshold $\tau_{\text{area}}$;
    \item shape IoU (after alignment) above $\tau_{\text{IoU}}$.
\end{itemize}

The subsequent steps, the moving sketch extraction and instruction generation, follow the same procedures described in the main paper (Sec. 3.2), using SEA-RAFT~\cite{10.1007/978-3-031-72667-5_3} for optical flow computation and Qwen3-VL for prompt generation, respectively. The final dataset comprises source images, target images, moving sketches, and corresponding instructions.

\section{More Data Examples}
We further present visual examples from our constructed dataset in Figure~\ref{fig:sup-data-show}. These examples illustrate the data structure, which includes the original image, the sketch condition overlaid on it, the corresponding target image with localized geometric modifications, and the VLM-generated text prompt.

\section{Additional Experimental Results}
In this section, we present supplementary experiments and qualitative results to further validate the effectiveness of our proposed method and its individual components.

\subsection{Robustness to Diverse Textual Prompts}
\label{sec:sup-prompt-generation}
We present qualitative results to demonstrate our method's robustness and flexibility when guided by a wide range of textual instructions. As illustrated in Figure~\ref{fig:sup-generation-prompt}, given a single source image and a fixed hand-drawn sketch that specifies the desired deformation of the mountain peak, we evaluate the model using prompts that differ in length and semantic complexity:
\begin{itemize}[label=(\alph*)]
    \item[(a)] \textit{$\langle sk \rangle$ Stretch the central mountain peak upward.}
    \item[(b)] \textit{$\langle sk \rangle$ Extend the highest snowy summit vertically to form a smooth arch.}
    \item[(c)] \textit{$\langle sk \rangle$ Deform the sharp mountain top into a tall, rounded curve.}
    \item[(d)] \textit{$\langle sk \rangle$ Elevate the central mountain peak, while strictly ensuring the clear blue sky and the dark forest in the foreground remain completely unaltered.}
\end{itemize}

Despite the differences in prompt length and semantic complexity, our approach consistently adheres to the structural guidance of the sketch. It successfully executes the local geometric modifications while preserving the unedited regions (e.g., the sky and non-target mountain areas).

\subsection{Qualitative Results on SPE Strategy}
We provide a qualitative result of the SPE strategy in Figure~\ref{fig:sup-spe}. Without SPE, the model lacks explicit pixel-level alignment between the source image and the sketch, leading to noticeable artifacts: the stretched tree shows hollow boundaries, and the enlarged rock appears unnaturally dark. With SPE,  these artifacts are largely alleviated, producing coherent and photorealistic local deformations.

\subsection{Qualitative Results on $\langle sk \rangle$ Position}
Figure~\ref{fig:sup-sk-position} provides a visual comparison of the learnable task-trigger token $\langle sk \rangle$ placed at the BOS versus the EOS. 
As illustrated in these examples, placing the token at the BOS better preserves the source content while achieving the intended edit, whereas placing it at the EOS can lead to noticeable content loss. For instance, the flower body nearly disappears in the first row, while several adjacent sea stacks disappear in the second row. In comparison, the BOS configuration produces more faithful local deformations and better preserves the unedited regions in these examples.

\subsection{Generalization to Hand-Drawn Sketches}
\begin{figure*}[!t]
  \centering
  \includegraphics[width=\linewidth]{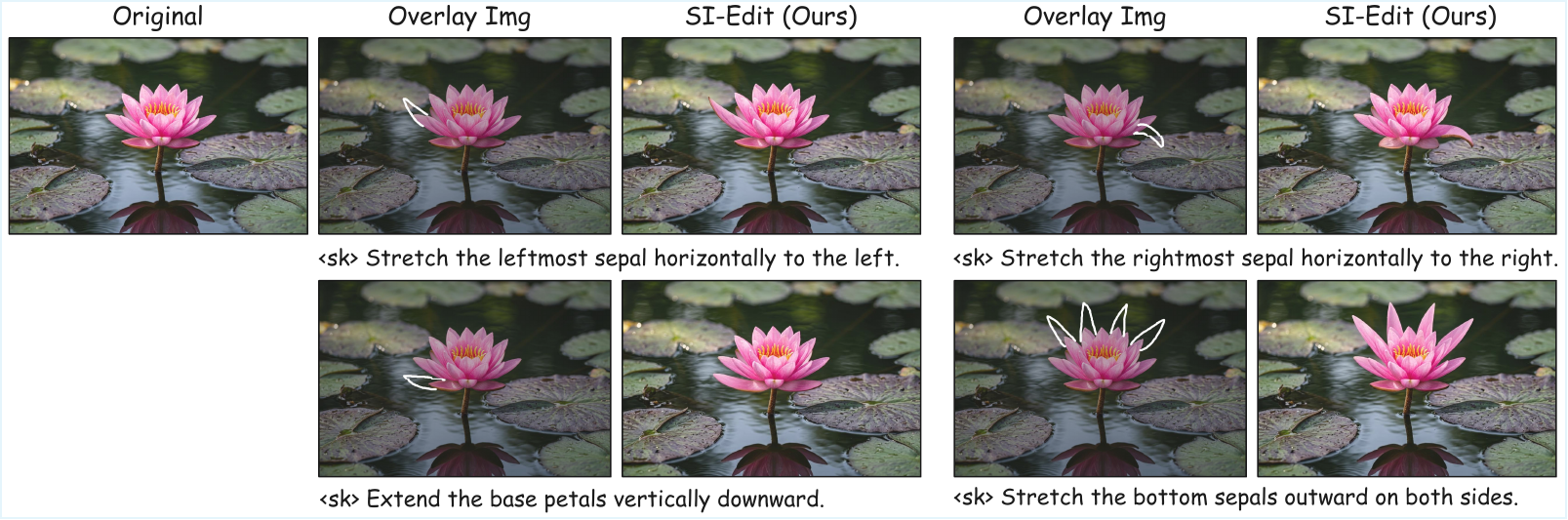}
  \vspace{-7mm}
  \caption{Generalization to hand-drawn sketches. For the same source image, each hand-drawn sketch specifies a different local deformation (e.g., stretching a specific sepal or petal). Our model consistently executes the edits while preserving the background and other unchanged elements.}
 \label{fig:sup-generation-sketch}
\end{figure*}

\begin{figure*}[!t]
  \centering
  \includegraphics[width=0.73\linewidth]{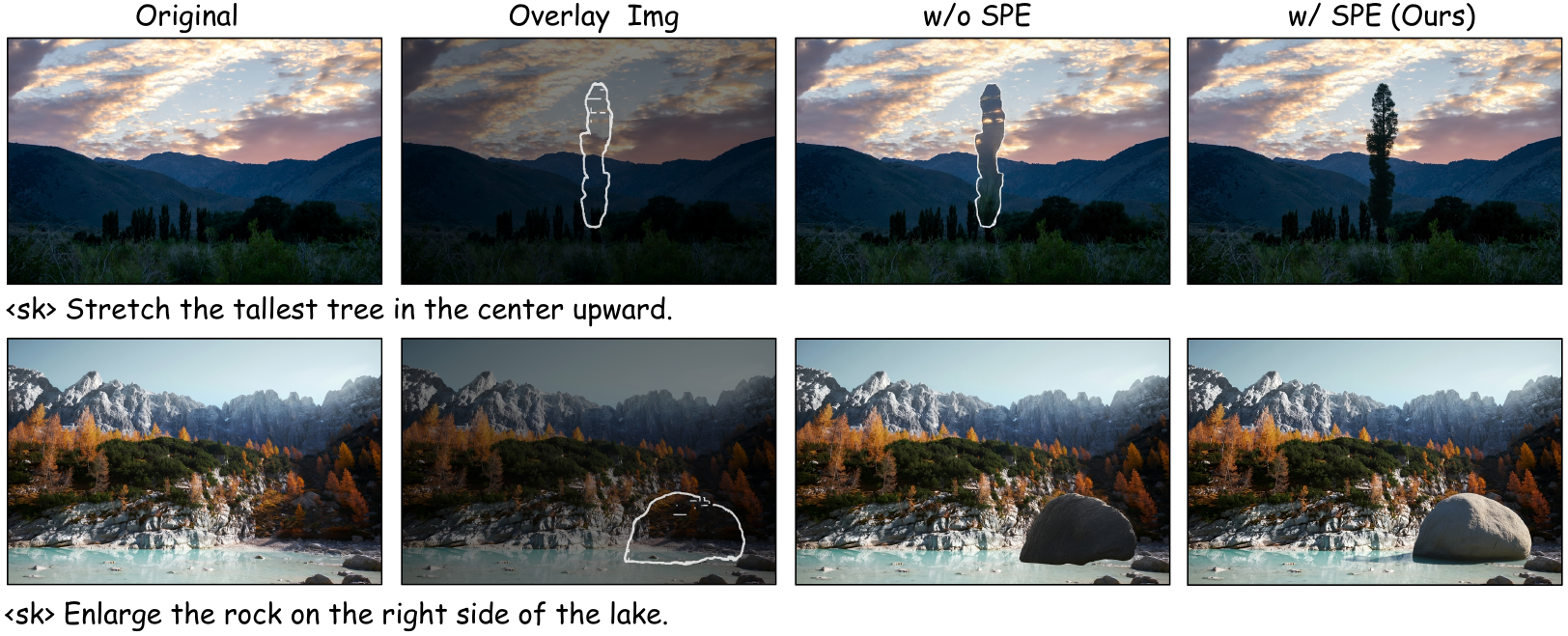}
  \vspace{-3mm}
  \caption{Qualitative results of the SPE strategy. Without (w/o) SPE, the model produces artifacts such as hollow boundaries in the stretched tree and unnatural darkening of the enlarged rock. With (w/) SPE, the edits blend naturally into the scene.}
 \label{fig:sup-spe}
\end{figure*}

\begin{figure*}[!t]
  \centering
  \includegraphics[width=0.73\linewidth]{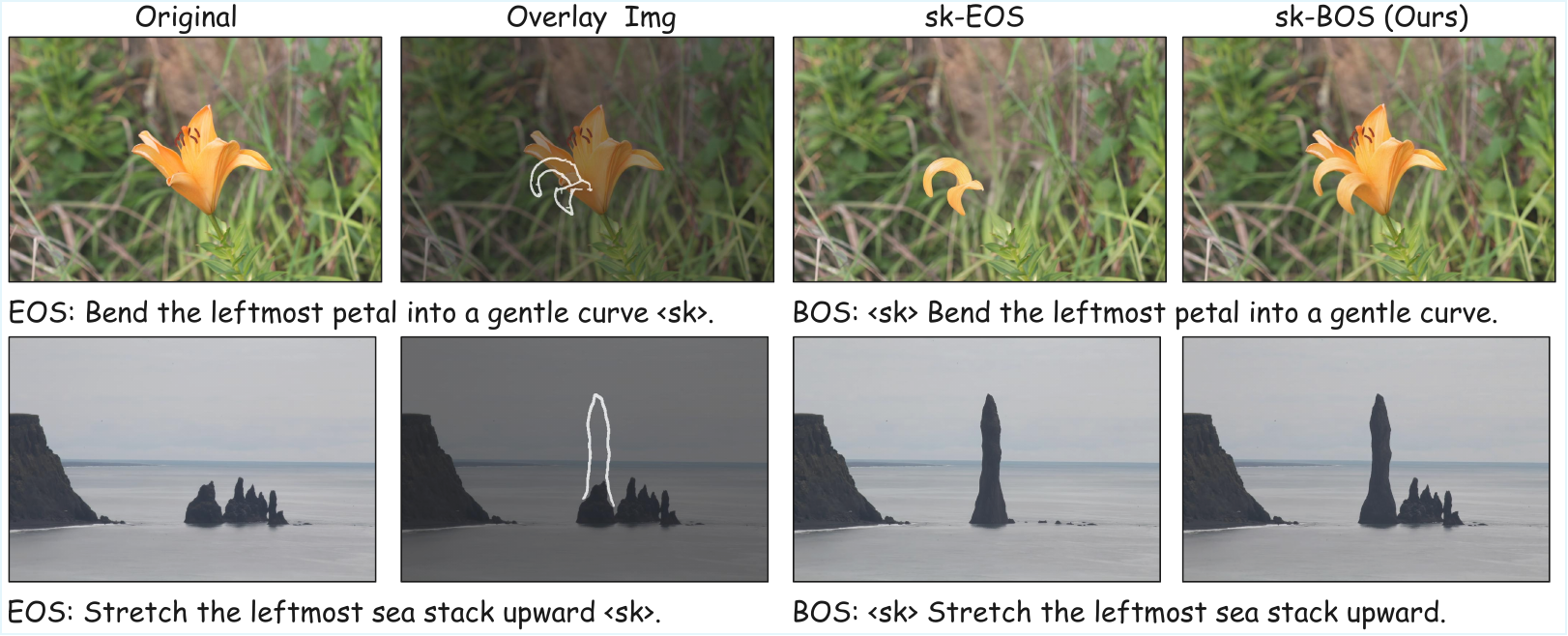}
  \vspace{-3mm}
  \caption{Qualitative results on the position of the task-trigger token $\langle sk \rangle$. Placing the token at the beginning of the instruction (sk-BOS) ensures accurate local geometric edits while maintaining high fidelity to the unedited regions. Conversely, placing it at the end (sk-EOS) leads to severe artifacts and loss of the original image content (e.g., the nearly missing flower body or disappearing adjacent sea stacks).}
 \label{fig:sup-sk-position}
\end{figure*}

We provide additional qualitative results in Figure~\ref{fig:sup-generation-sketch}. For the same source image, we apply multiple hand-drawn sketches specifying different local deformations (e.g., stretching or extending specific petals in different directions). The results show that the model follows the given sketches while preserving the unedited background and overall image structure.

\subsection{User Study Details}
\label{sec:sup-user-study}
\begin{figure}[!t]
  \centering
  \includegraphics[width=0.96\linewidth]{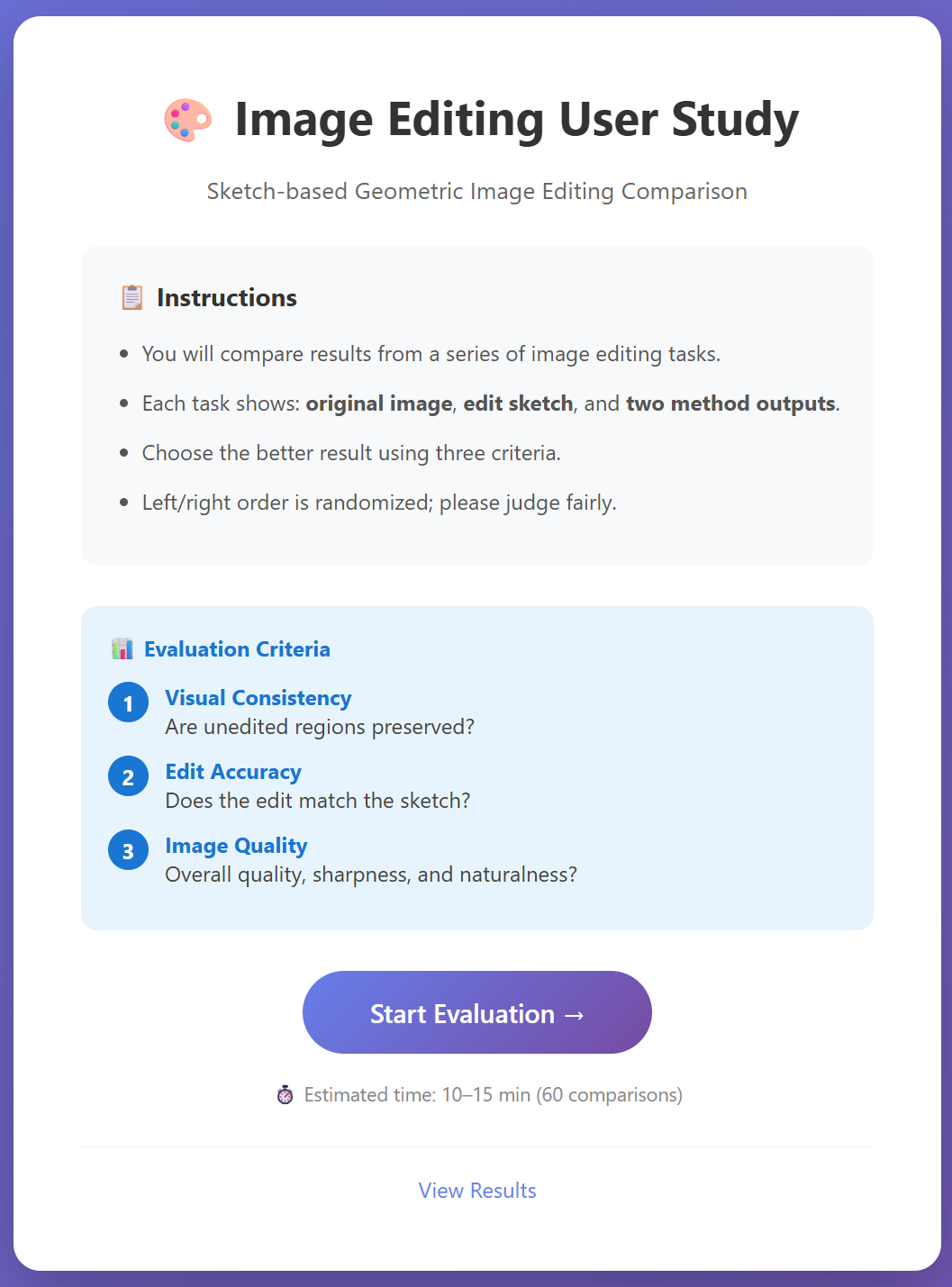}
  \vspace{-2mm}
  \caption{
  User study start page. 
  }
  \label{fig:user-study-instructions}
\end{figure}

\begin{figure}[!t]
  \centering
  \includegraphics[width=0.96\linewidth]{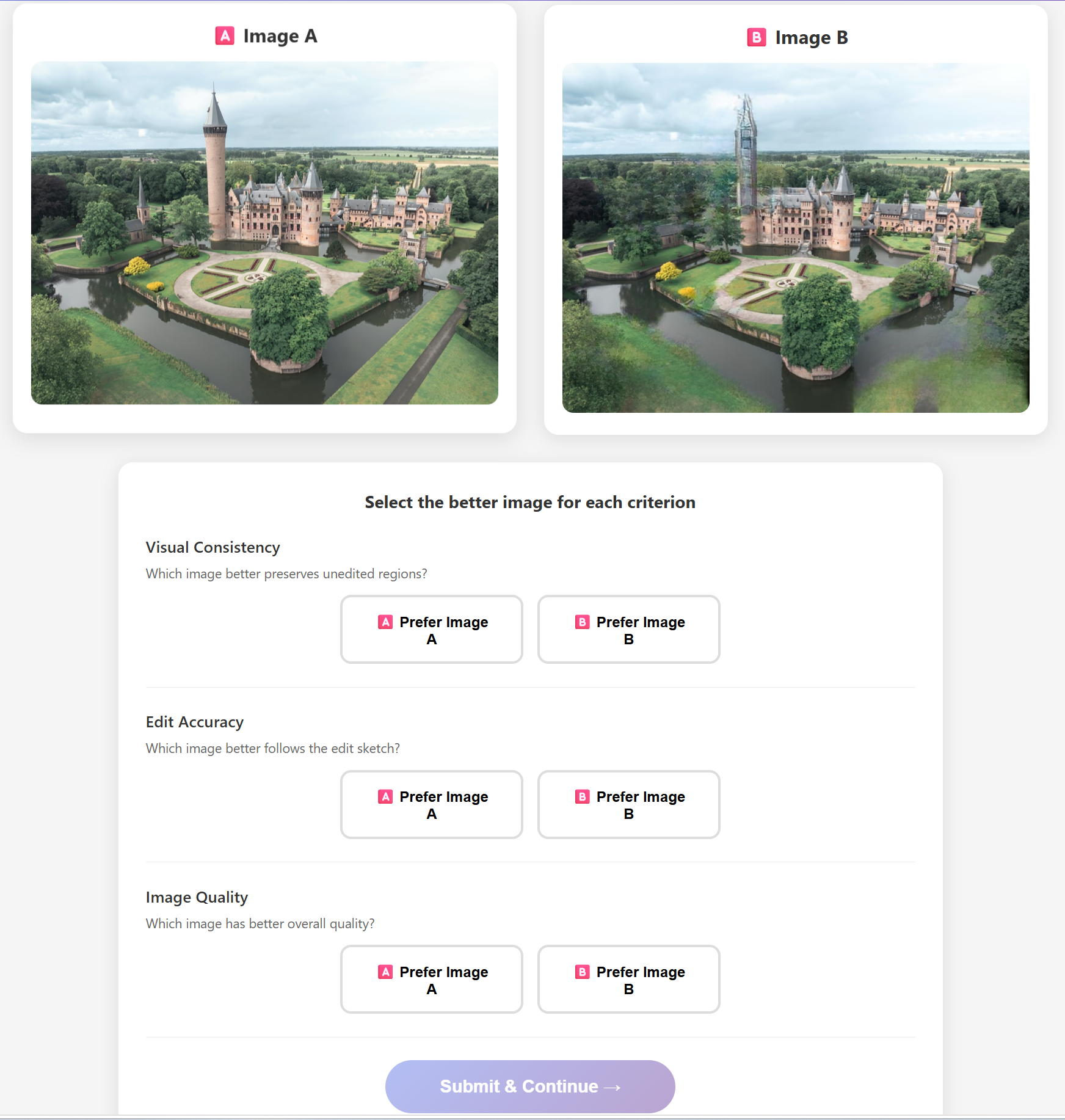}
  \caption{Example of the pairwise comparison interface. 
  }
  \label{fig:user-study-interface}
\end{figure}

\begin{figure}[!t]
  \centering
  \includegraphics[width=1.0\linewidth]{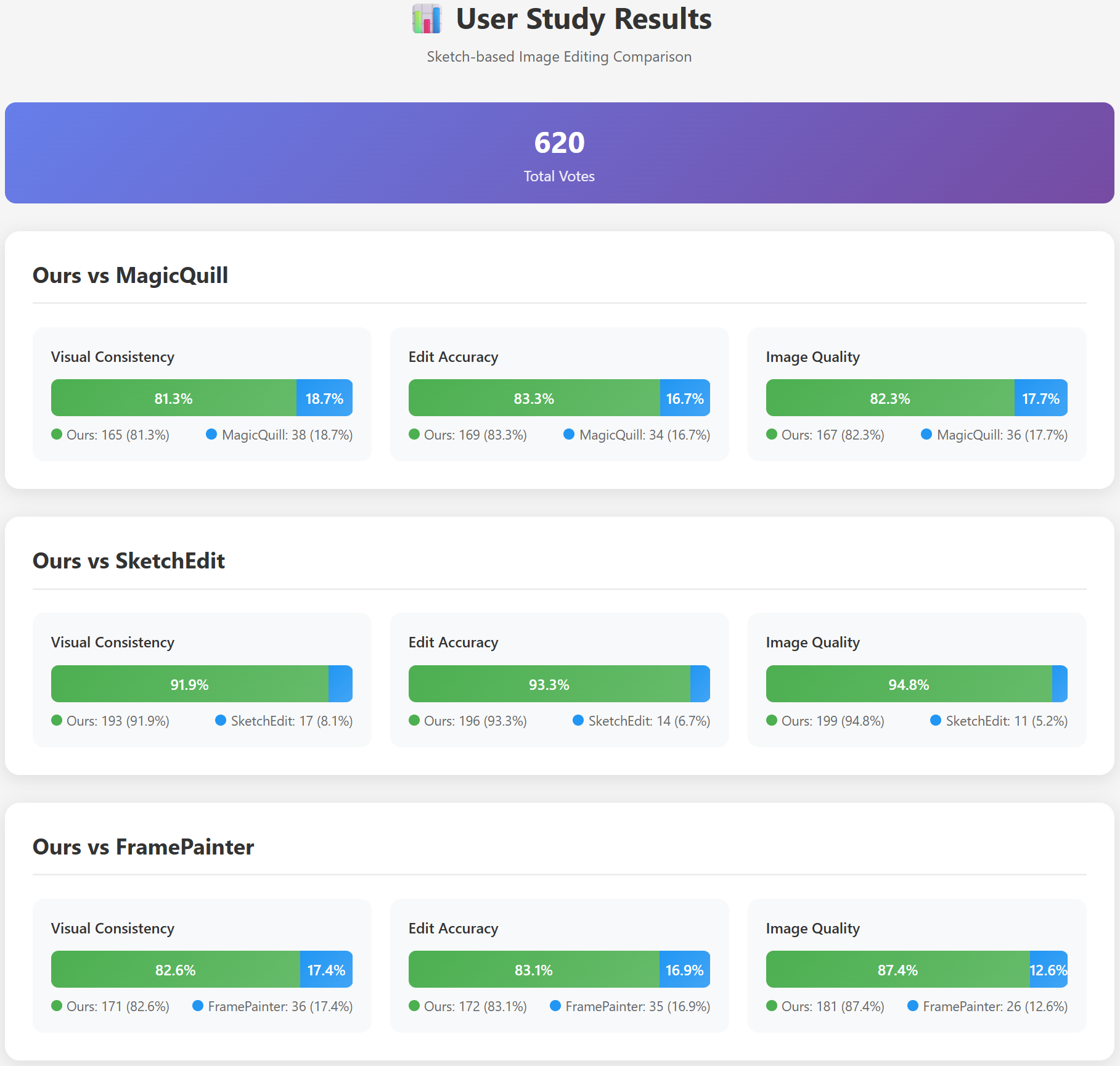}
  \caption{
  Aggregated results.
  The bars report the preference rates for SI-Edit over MagicQuill, SketchEdit, and FramePainter in terms of visual consistency, edit accuracy, and image quality.
  }
  \label{fig:user-study-results}
\end{figure}

To further assess the practical usability of our method, we conduct a user study with human participants.
We conduct a pairwise user study on 60 test cases randomly selected from our 159-sample test set. As illustrated in Figure~\ref{fig:user-study-instructions}, participants evaluate the editing results according to three criteria: visual consistency, edit accuracy, and overall image quality. For each comparison, two anonymized results are presented in a randomized left--right order, and participants independently select their preferred result for each criterion. An example of the evaluation interface is shown in Figure~\ref{fig:user-study-interface}.
In total, we obtain 620 pairwise comparisons (1,860 criterion-specific judgments).
Figure~\ref{fig:user-study-results} summarizes the percentage of comparisons in which SI-Edit is preferred over MagicQuill, SketchEdit, and FramePainter. The results show that SI-Edit receives consistently higher preference rates than all three baselines across the three evaluation criteria.

\end{document}